\documentclass[3p, times]{elsarticle}

\usepackage{lineno,hyperref}
\usepackage{amsmath,amssymb}
\usepackage{subfigure}
\usepackage{multirow}
\usepackage{array}
\usepackage{algorithm}
\usepackage{algorithmic}

\makeatletter
\newcommand{\thickhline}{%
    \noalign {\ifnum 0=`}\fi \hrule height 1pt
    \futurelet \reserved@a \@xhline
}
\newcommand{\topcaption}{%
\setlength{\abovecaptionskip}{0pt}%
\setlength{\belowcaptionskip}{8pt}%
\caption}

\modulolinenumbers[5]

\journal{Journal of \LaTeX\ Templates}

\begin{document}

\begin{frontmatter}

\title{Detecting Multi-Oriented Text with Corner-based Region Proposals}

\author[a]{Linjie Deng}
\author[a]{Yanxiang Gong}
\author[b]{Yi Lin}
\author[a]{Jingwen Shuai}
\author[a]{Xiaoguang Tu}
\author[c]{Yuefei Zhang}
\author[a]{Zheng Ma}
\author[a]{Mei Xie\corref{ca}}

\address[a]{School of Information and Communication Engineering, UESTC, Chengdu, China}
\address[b]{National Key Laboratory of Fundamental Science on Synthetic Vision, Sichuan University, Chengdu, China}
\address[c]{Chongqing Institute of Public Security Science and Technology, Chongqing, China}

\cortext[ca]{Corresponding author}
\ead{mxie@uestc.edu.cn}

\begin{abstract}
Previous approaches for scene text detection usually rely on manually defined sliding windows. This work presents an intuitive two-stage region-based method to detect multi-oriented text without any prior knowledge regarding the textual shape. In the first stage, we estimate the possible locations of text instances by detecting and linking corners instead of shifting a set of default anchors. The quadrilateral proposals are geometry adaptive, which allows our method to cope with various text aspect ratios and orientations. In the second stage, we design a new pooling layer named Dual-RoI Pooling which embeds data augmentation inside the region-wise subnetwork for more robust classification and regression over these proposals. Experimental results on public benchmarks confirm that the proposed method is capable of achieving comparable performance with state-of-the-art methods. The code is publicly available at \url{https://github.com/xhzdeng/crpn}.

\end{abstract}

\begin{keyword}
Multi-Oriented Text Detection\sep Dual-RoI Pooling\sep Corner-based Region Proposal Network
\end{keyword}

\end{frontmatter}


\section{Introduction}
Automatically reading text in the wild is a fundamental problem of computer vision since text in scene images commonly conveys valuable information. It has been widely used in various applications such as multilingual translation, automotive assistance and image retrieval. Normally, a reading system consists of two sub-tasks: detection and recognition. This work focuses on text detection, which is the essential prerequisite of the subsequent processes in the whole workflow.

Though extensively studied \cite{SWT2010CVPR, RTTLR2012CVPR, FASTEXT2015ICCV,RTCNN2016IJCV} in recent years, scene text detection is still enormously challenging due to the diversity of text instances and undesirable image quality. Before the era of deep learning \cite{ALEXNET2012NIPS}, most related works utilized the sliding window \cite{RTCNN2016IJCV, STLD2015CVPR} or connected component \cite{SWT2010CVPR, RTTLR2012CVPR, FASTEXT2015ICCV} with hand-crafted features. The methods based on the sliding window detect texts by shifting several windows onto all positions in multiple scales. Although these methods are able to achieve a high recall rate, the large number of candidates may result in low precision and heavy computations. The approaches based on the connected component focus on the detection of individual characters and the relationships between them. Although these methods are faster than previous ones, errors will occur and accumulate throughout each of the sequential steps \cite{TF2015ICCV}, which may degenerate the performance of detection.

Recently, benefited from the significant achievements of generic object detection based on deep neural network \cite{CNN1998}, methods with high performance \cite{FASTERRCNN2015NIPS, YOLO2016CVPR, SSD2016ECCV} have been modified to detect horizontal scene text \cite{DEEPTEXT2016ARXIV, FCRN2016CVPR, TEXTBOX2017AAAI} and the results have amply demonstrated their effectiveness. In addition, in order to achieve multi-oriented text detection, some methods \cite{RRPN2017ARXIV, DMPN2017CVPR} designed several rotated anchors to find the best matched proposals to inclined text instances. Although these methods have shown their promising performances, as \cite{DMPN2017CVPR} refers, the strategy of man-made shapes of anchors may not be the optimal designs. 

\begin{figure}[!htb]
  \centering
  \subfigure[]{
    \label{Fig_1_a} 
    \includegraphics[height=3.0cm]{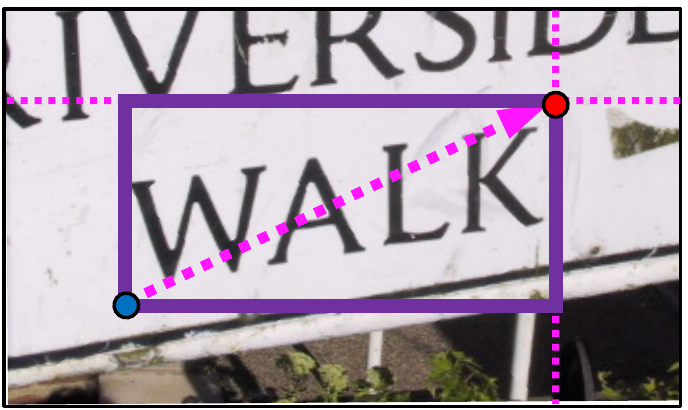}}
  \hspace{-0.5ex}
  \subfigure[]{
    \label{Fig_1_b} 
    \includegraphics[height=3.0cm]{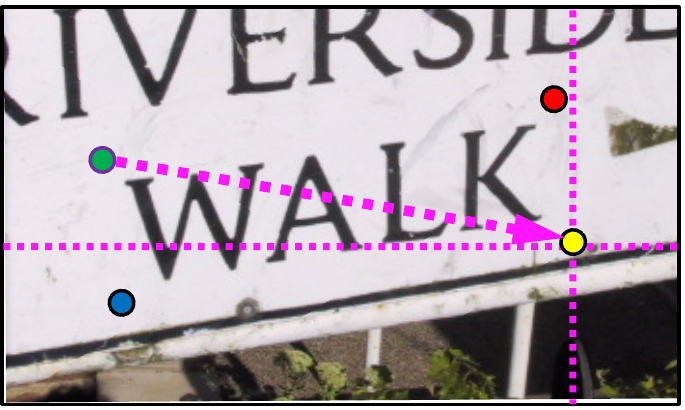}}
  \hspace{-0.5ex}
  \subfigure[]{
    \label{Fig_1_c} 
    \includegraphics[height=3.0cm]{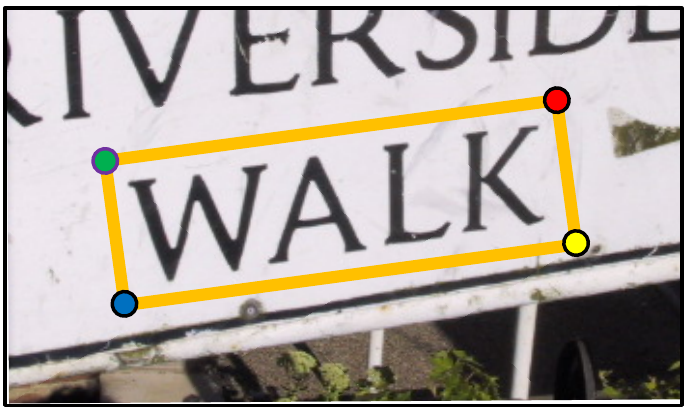}}
  \caption{\textbf{Annotating an instance of text:} : (a) The conventional way of drawing a rectangular bounding-box. Clicking on a corner (blue point) and dragging the mouse to the diagonally opposite corner (red point) to receive a rectangle (solid purple box). But the rectangular bounding-box is not very accurate for rotated text region. (b) (c) Drawing another diagonal (dashed arrow in (b)) to finish the annotation (solid orange box in (c)) for inclined text instance. Best viewed in color.}
  \label{Fig_1}
\end{figure}

In this paper, we tackle scene text detection with a new perspective which is mostly learned from the process of object annotation. As shown in Fig.\ref{Fig_1}, the most common way \cite{EC2017ICCV, DEC2017ARXIV} to make an annotation is to click on a corner of an imaginary rectangle enclosing the object, and then drag the mouse to the diagonally opposite corner to get a rectangular box. To annotate the inclined text instance, we need another diagonal to obtain a quadrilateral bounding-box. If the box is not particularly accurate, users can make further amendments by adjusting the coordinates of the corners. Following the above procedure, our motivation is to harness the corners to infer the locations of text bounding-boxes. Basically, the proposed method is a two-stage region-based \cite{RCNN2014CVPR} detection framework. In region proposal stage, we abandon the anchor fashion and utilize corners to generate quadrilateral proposals for estimating the possible locations of text instances. In region classification stage, a region-wise network is trained for text/non-text classification and corner regression over these proposals. Based on corners, the quadrilateral outputs are flexible to capture various text aspect ratios and orientations. The resulting model is trained end-to-end and obtains an F-measure 0.876 on ICDAR 2013, 0.845 on ICDAR 2015 and 0.591 on COCO-Text. Besides, compared with recent published works \cite{DDR2017ICCV, SSTD2017ICCV}, it is competitively faster in running speed.

We summarize our primary contributions as follows:

\begin{enumerate}[(1)]
\item Dissimilar to previous works heavily relying on the well-designed anchors, our new corner-based region proposal network generates quadrilateral region proposals by detecting and linking the corners of text bounding-boxes.
\item In order to suppress the negative linkages, we design a new variable named Link Direction to indicate where to find the partner for each detected corner candidate.
\item A nearly cost-free module named Dual-RoI Pooling which embeds data augmentation inside the region-wise subnetwork is presented to improve the utilization of training data.
\item Our scene text detector achieves strong performances and competitive speeds on public benchmarks. All of our training and testing code is open source now.
\end{enumerate}

\section{Methodology}
In this section, we will describe the proposed method. Technically, our method is based on Faster R-CNN \cite{FASTERRCNN2015NIPS} and DeNet \cite{DENET2017ARXIV} detection frameworks. We make some innovations to combine and extend them for detecting multi-oriented text. Details will be delineated in the following.

\subsection{Corner-based Region Proposal Network}
Here we introduce our new region proposal algorithm named Corner-based Region Proposal Network (CRPN). It draws primarily on DeNet \cite{DENET2017ARXIV}, which is a novel generic object detection framework. We extend the Corner-based RoI Detector which can only output rectangular proposals in DeNet to generate quadrilateral ones for matching arbitrary-oriented text instances. In general, two intersected diagonals can determine a quadrangle. The main purpose of CRPN is to search corners on the whole image and find the intersected diagonals which link them. It mainly consists of two steps:  corner detection and proposal sampling.

\subsubsection{Corner Detection}
Similar to semantic segmentation, the task of corner detection is performed by predicting the probability of each position $(x,y)$ that belongs to one of the predefined corner types. Owing that texts are usually very close to each other in natural scenes, it is hard to separate corners which are from adjacent text instances. Thus, we use the one-versus-rest strategy and apply four-branches ConvNet to predict the independent probability denoted as $P_i(x,y)$ for each corner type, where $i\in\{top\_left,top\_right,bot\_right,bot\_left\}$. Obviously, only the $top\_left$ and $bot\_right$ or $top\_right$ and $bot\_left$ can be connected diagonally. However, most of the linkage is negative because there is only one positive partner for each corner candidate. For the purpose of suppressing the negative linkages, we define a new variable named Link Direction. Let $\theta(p,q)$ be the orientation of any diagonal $\overrightarrow{pq}$ to the horizontal, $p$ and $q$ are two corner candidates from two types which can be linked into a diagonal. Then $\theta(p,q)$ is discretized into one of $K$ values, as HOG \cite{HOG2005CVPR} did. The Link Direction of corner $p$ to $q$ is calculated as follow:

\begin{gather}
D_{\overrightarrow{pq}}=ceil\left(\frac{K\ast\theta(p,q)}{2\pi}\right)
\label{Eq:LinkDirection}
\end{gather}

\noindent Then we convert the binary classification problem (corner/non-corner) to a multi-class one, where each class corresponds to a value of Link Direction. To this end, each corner detector outputs a prediction map with channel dimension $K+1$ (plus one for background and other corner types). The independent probability of each position which belongs to corner type $i$ is given by:

\begin{gather}
P_i(x,y)=\sum_{k=1}^{K}P^{'}_i(k|x,y), k\in\{0,1,\ldots,K\}
\label{Eq:pointscore}
\end{gather}

\begin{figure}[!htb]
  \centering
  \subfigure[]{
    \label{Fig_2_a} 
    \includegraphics[height=2.5cm]{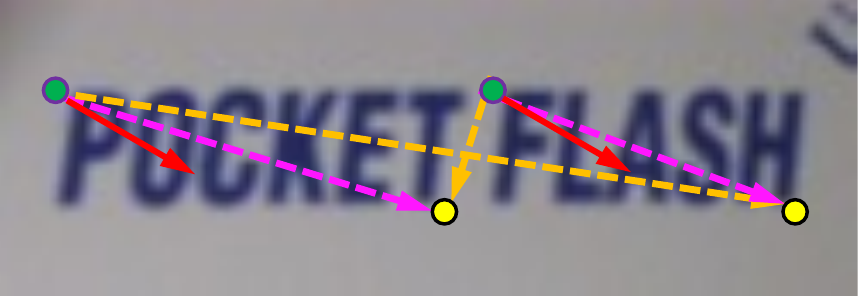}}
  \hspace{-0.5ex}
  \subfigure[]{
    \label{Fig_2_b} 
    \includegraphics[height=2.5cm]{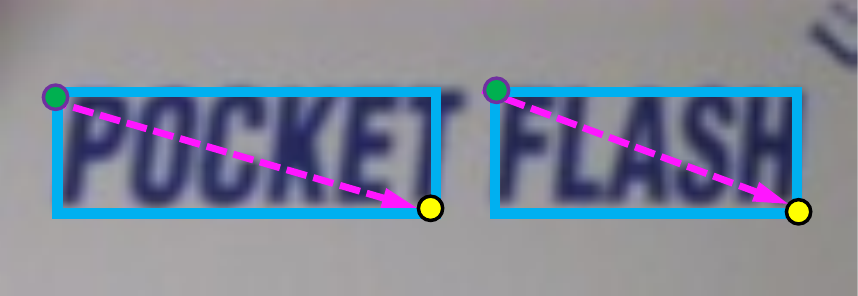}}
  \caption{\textbf{Suppressing the negative linkages via Link Direction:} The green and yellow points are two types of detected corners which can be connected diagonally. (a) The solid (red) and dashed (orange and purple) arrows represent the practical and predicted Link Direction respectively. The dashed orange arrows will be filter out via Eq.\ref{Eq:LinkFilter}. (b) The output region proposals (solid blue box) are generated by dashed purple arrows. Best viewed in color.}
  \label{Fig_2} 
\end{figure}

\noindent $P^{'}_i(k|x,y)$ is computed by a softmax over the $K+1$ output maps of corner class $i$. As \cite{SWT2010CVPR} applying the gradient direction $d_p$ of each edge pixel $p$ to find another edge pixel $q$ where the gradient direction $d_q$ is roughly opposite to $d_p$, we filter out the linkages which:

\begin{gather}
|D_{\overrightarrow{pq}}-D_p|>1
\label{Eq:LinkFilter}
\end{gather}

\noindent where $D_p$ is the predicted Link Direction of corner $p$, $D_{\overrightarrow{pq}}$ is the practical Link Direction of corner $p$ to $q$ which can be easily calculated via Eq.\ref{Eq:LinkDirection}. As depicted in Fig.\ref{Fig_2}, the Link Direction can indicate where to find the partner for each corner, and meanwhile is helpful for separating two adjacent texts.

\subsubsection{Proposal Sampling}
We develop a simple algorithm for searching corner candidates from probability maps and assembling them into quadrilateral proposals. Specially, in order to improve the recall rate, we only use three different types of corners to generate a proposal. The working steps of algorithm are described as follows:

\begin{algorithm}
	\renewcommand{\algorithmicrequire}{\textbf{Input:}}
	\renewcommand{\algorithmicensure}{\textbf{Output:}}
	\caption{Generating Proposals by Searching and Linking Corners}
	\label{Alg_1}
	\begin{algorithmic}
		\REQUIRE the outputs of each corner detector
		\ENSURE quadrilateral proposals
		\STATE \textbf{Step 1:} Predefine a threshold $T$ and search the probability map for corner candidates where $P_i(x,y) > T$. A candidate which is close to higher scoring selected one will be rejected and only top-$M$ ranked candidates will be used in next step. 
		\STATE \textbf{Step 2:} Generate a set of unique diagonals by linking every candidate of $top\_left$ and every one of $bot\_right$. The negative linkages will be filtered out via Eq.\ref{Eq:LinkFilter}. 
		\STATE \textbf{Step 3:} Generate another set of diagonals by rotating each diagonal to collinear with every corner of last two types. A quadrilateral proposal will be generated by two intersected diagonals (source one and rotated one). 
		\STATE \textbf{Step 4:} Repeat \textbf{Step 2} and \textbf{Step 3} with corners of types $top\_right$ and $bot\_left$.
	\end{algorithmic}  
\end{algorithm}


With the probability of corner, we can easily estimate the score that a proposal $B$ contains a text instance by applying a Na{\"\i}ve Bayesian Classifier to each corner of $B$:

\begin{gather}
P(B)=\prod_{i}P_i(x_i,y_i)
\label{Eq:BoxScore}
\end{gather}

\noindent where $(x_i,y_i)$ indicates the coordinates of the corner type $i$ associate with proposal $B$. We also adopt non-maximum suppression (NMS) on proposals based on their score to reduce redundancy because most of the proposals highly overlap with each other. Considering the computational accuracy of the standard NMS based on the IoU of rectangles is unsatisfactory for quadrilateral proposals, we use the algorithm introduced by \cite{RRPN2017ARXIV}, which can compute the IoU of quadrangles, to solve this problem. Then the vast majority of proposals will be discarded after this operation.

\subsection{Dual-RoI Pooling}

\begin{figure}[!htb]
\centering
\includegraphics[height=7.0cm]{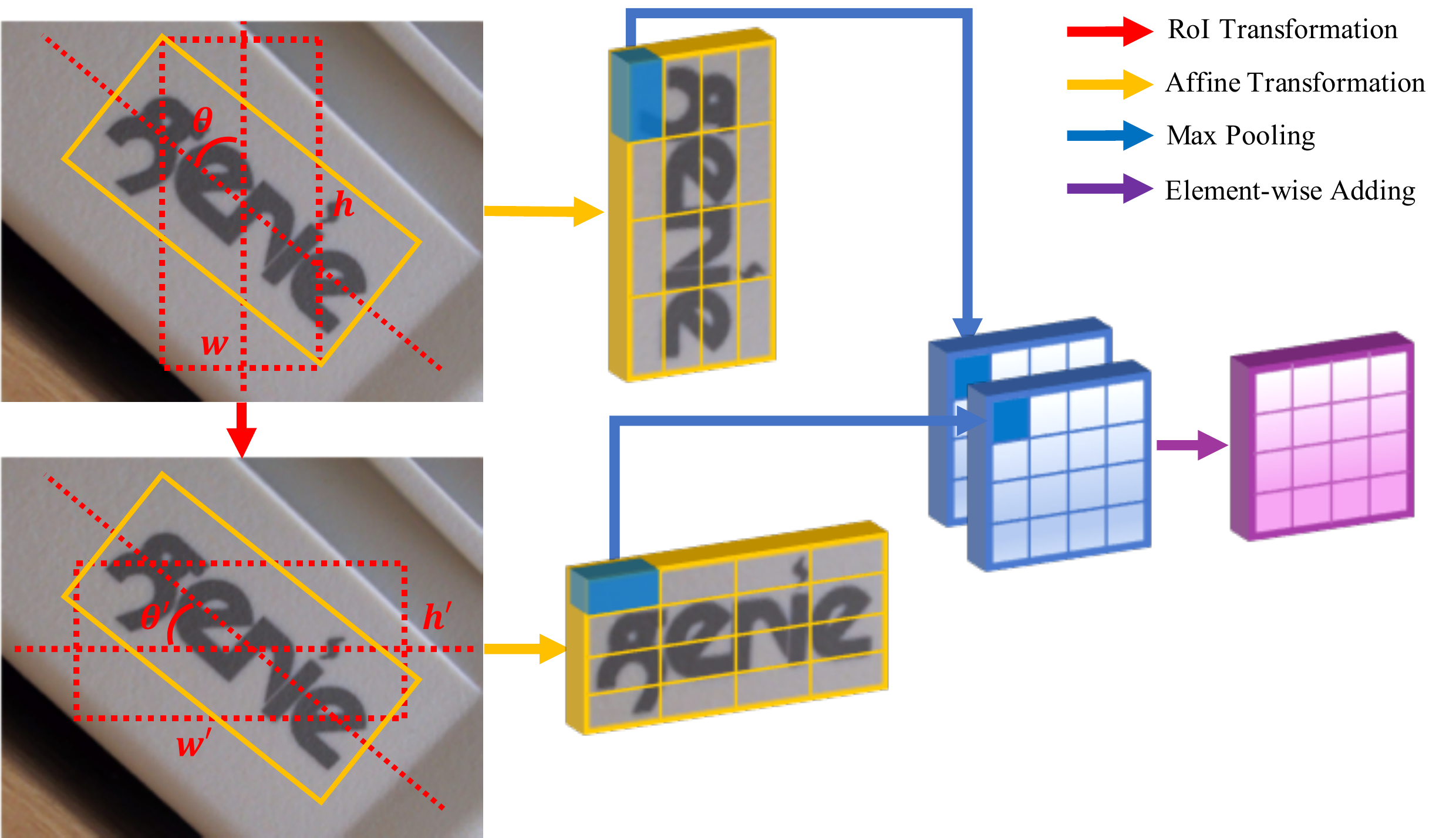}
\caption{\textbf{Dual-RoI Pooling:} Actually, the Dual-RoI Pooling is performed on the feature map. We draw it on source image for better viewing. Different colored arrows represent different operations. Best viewed in color.}
\label{Fig_3}
\end{figure}

As presented in \cite{FASTRCNN2015ICCV}, the RoI pooling layer extracts a fixed-length feature vector from the feature map for each region of interest (RoI), and then each vector is fed into a sequence of fully connected layers for further classification and bounding-box regression. In our task, the traditional RoI pooling which can only handle axis-aligned RoIs is also not accurate for quadrilateral ones. Thus, the Rotation RoI (RRoI) pooling presented by \cite{RRPN2017ARXIV} is adopted to address this issue. Each RRoI is represented by a five-tuple $(r,c,h,w,\theta)$ where $r,c$ are coordinates of bounding-box center, $h,w$ are height and width of bounding-box, and $\theta$ is the rotation angle to the horizontal. This operation works by mapping the rotated RoI into an axis-aligned one via affine transformation, and then dividing the $h\times{w}$ window into an $H\times{W}$ grid of sub-windows and max-pooling the values in each sub-window into the corresponding output grid cell, as shown in Fig.\ref{Fig_3}. Since the input format of RRoI pooling is rotated rectangle, we convert the quadrilateral proposals into rotated rectangular ones through method $minAreaRect$ in OpenCV and $\theta$ is within the interval $[0^{\circ}, 90^{\circ})$.

Usually, for improving the robustness of the model when encountering various orientations of texts, existing methods \cite{DDR2017ICCV, FTSN2017ARXIV, RRCNN2017ARXIV} make use of data augmentation that rotates source images to different angles to harvest sufficient data for training. Despite the effectiveness of data augmentation, the main drawback lies in learning all the possible transformations of augmented data require more network parameters, and it also may result in significant increase of training cost and over-fitting risk \cite{ORN2017CVPR}. With the aim to alleviate these drawbacks, we present a built-in data augmentation named Dual-RoI Pooling. For each RRoI $O$ which represented by $(r,c,h,w,\theta)$, we transform it into another RRoI $O^{'}$ represented by $(r^{'},c^{'},h^{'},w^{'},\theta^{'})$ where:

\begin{gather}
r^{'}=r,\quad c^{'}=c,\quad h^{'}=w,\quad w^{'}=h,\quad \theta^{'}=90^{\circ}-\theta
\label{Eq:RoITrans}
\end{gather}

\noindent As described in Fig.\ref{Fig_3}, inputting the RoI $O^{'}$ to RRoI pooling layer will obtain a totally different grid of sub-windows. We combine these two RoIs as a Dual-RoI and fuse their feature vectors to get the final output via element-wise adding operation. The essence of Dual-RoI pooling embeds multi-instance learning \cite{DML2015CVPR} inside the region-wise subnetwork and that will be helpful for finding the most representative features for training, similar to TI-POOLING \cite{TIPOOL2016CVPR}. As a result of directly conducting the transformation on feature map instead of on the source image, our module is more efficient than \cite{TIPOOL2016CVPR}. Considering the diversity of text arrangement, we argue that the element-wise adding is more appropriate than element-wise maximum in our task.

\subsection{Network Architecture}
\begin{figure}[!htb]
\centering
\includegraphics[height=5.5cm]{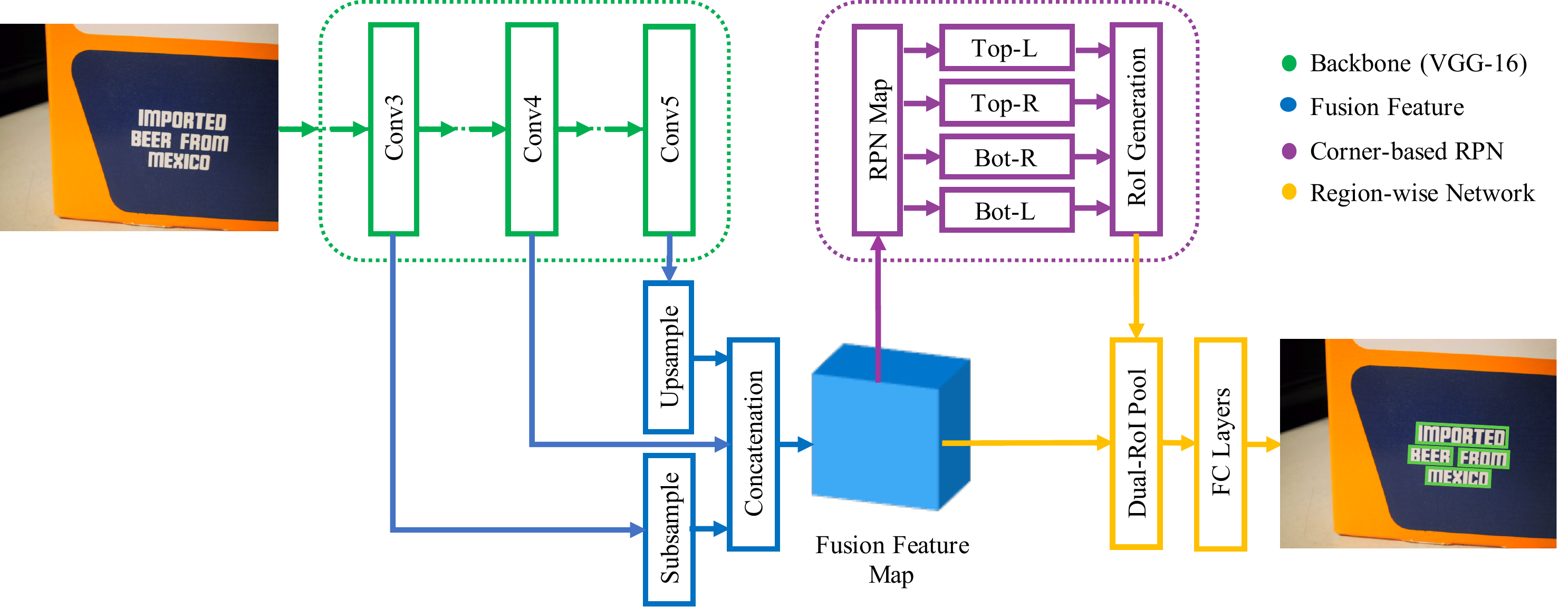}
\caption{\textbf{The architecture of the proposed detection network.} Best viewed in color.}
\label{Fig_4}
\end{figure}

The network architecture of the proposed method is diagrammed in Fig.\ref{Fig_4}. All of our experiments are implemented on VGG-16 \cite{VGG2015ICLR}, though other networks are also applicable. To obtain more accurate corners prediction, we choose the output of $Conv4$ as the final feature map, whose size is $1/8$ of the input image. Which has been proven in many works \cite{HYPERNET2016CVPR, PVANET2016ARXIV}, combining the fine-grained details from low level layers with coarse semantic information from high level layers will be helpful for the detector. Our structure of feature fusion mainly inherits from PVANet \cite{PVANET2016ARXIV} but is slightly different to it. We add a deconvolutional layer to carry out upsampling, and a ${3}\times{3}\times{64}$ convolutional layer with $stride=2$ to conduct subsampling. The concatenated features are combined by a ${1}\times{1}\times{512}$ convolutional layer.

\subsection{Loss Function}
Based on the above definitions, we give the multi-task loss $L$ of proposed network. The model is trained to simultaneously minimize the losses on corner segmentation, proposal classification and corner regression. Overall, the loss function is a weighted sum of these losses:

\begin{align}
L=\lambda_{seg}L_{seg}+\lambda_{cls}L_{cls}+\lambda_{loc}L_{loc}
\label{Eq:Loss_ALL}
\end{align}

\noindent where $\lambda_{seg}, \lambda_{cls}, \lambda_{loc}$ are user defined constants indicating the relative strength of each component. In our experiment, all weight constants $\lambda$ are set to 1.

\subsubsection{Loss for Segmentation}
In training stage, the loss function for segmentation task is computed over all pixels in the output map with the ground-truth which is identified by mapping the corners to a single position in the label map, corners out of boundary are simply discarded.  For a typical natural image, the distribution of corner/non-corner pixels is heavily biased. It is possible to optimize for the loss functions of all pixels, but this will bias towards negative samples as they are dominant. Instead, we randomly sample 32 pixels as a mini-batch which the samples positive and negative pixels have a ratio of up to 1:1. If the number of positive pixels is not enough, we pad the mini-batch with negative ones. Furthermore, we adopt a weighted softmax loss function introduced in \cite{FSDS2016CVPR} to balance the loss between the corner/non-corner classes, given by:

\begin{gather}
L_{seg}=\sum_{i}-\frac{1}{|S_i|}\sum_{m=1}^{|S_i|}\sum_{k=0}^{K}\omega_i^kI(z_i^m=k)logP_i^{'}(k|s_i^m)
\label{Eq:Loss_Seg}
\end{gather}

\noindent $S_i$ is a mini-batch of corner type i. $|S_i|$ is the number of samples. $K$ denotes the maximum value of the Link Direction. $z_i^m$ is the ground-truth of the $m$-th sample $s_i^m$. The function $I(\cdot)$ evaluates to 1 when $z_i^m=k$ and 0 otherwise. $P_i^{'}(k|s_i^m)$ represents how likely the value of Link Direction of $s_i^m$ is $k$. The $w_i^k$ is the balanced weight between corner and non-corner samples, given by:

\begin{align}
\omega_i^k = \left\{
             \begin{array}{ll}
             {\dfrac{\sum_{m}{I(z_i^m==0)}}{|S_i|}} & \text{if}\ k>0 \, \\
             {\dfrac{\sum_{m}{I(z_i^m>0)}}{|S_i|}} & \text{otherwise} \,
             \end{array}
           \right.
\label{Eq:Loss_Seg_Weight}
\end{align}

\subsubsection{Loss for Region-wise Network}
As in \cite{FASTRCNN2015ICCV}, the region-wise network has two sibling outputs: one for proposal classification and another for corner regression. The first objective aims for distinguish texts from non-texts with softmax loss, as follows:

\begin{gather}
L_{cls}=ylog(p)+(1-y)log(1-p)
\label{Eq:Loss_Cls}
\end{gather}

\noindent where $y$ equals 1 if input sample is text, otherwise is 0. The $p$ is text confidence score. The second objective attempts to regress the fine bounding-box, as follows:

\begin{gather}
L_{loc}=smooth_{L_1}(t-t^{*})
\label{Eq:Loss_Loc}
\end{gather}

\noindent $t$ and $t^{*}$ represents the predicted and target regression tuples respectively, and $smooth_{L_1}$ is a robust $L_1$ loss defined in R-CNN [5]. The regression target $t^{*}$ consists of eight terms:

\begin{gather}
t_{x_i}^{*}=(x_i-x_i^*)/w,\quad t_{y_i}^{*}=(y_i-y_i^*)/h
\label{Eq:Reg_Target}
\end{gather}

\noindent $(x_i,y_i)$ and $(x_i^*,y_i^*)$ denote the coordinates of the corner type $i$ for proposal and target bounding-box respectively. $w$ and $h$ are the width and height of minimal bounding rectangle of input proposal. The target bounding-box for each proposal is identified by selecting the ground-truth with the largest overlap.

\section{Experiments}
In the following part, we will describe the implementation details of our method. We also run a number of ablations to analyze the effectiveness of the proposed component. Finally, we will present the evaluation results on three public benchmarks: ICDAR 2013 \cite{ICDAR2013}, ICDAR 2015 \cite{ICDAR2015} and COCO-Text \cite{COCO2016ARXIV}.

\subsection{Implementation Details}
The backbone of network is initialized by pretraining a model for ImageNet \cite{IMAGENET2015IJCV} classification, and all other new layers are initialized by "Xavier" \cite{XAVIER2010AISTATS}. The training images are collected from SynthText \cite{FCRN2016CVPR}, ICDAR 2013 and ICDAR 2015. We randomly pick up 100,000 images from SynthText for pretraining, and then the real data from the training sets of ICDAR 2013 and 2015 is used to finetune a unified model. The model is trained end-to-end by using the standard SGD algorithm. Momentum and weight decay are set to 0.9 and $5\times10^{-4}$  respectively. Following the multi-scale training in \cite{PVANET2016ARXIV}, we resize the images in each training iteration such that the short side of input is randomly chose between 480 and 800. In pretraining stage, the learning rate is set to $10^{-3}$ for the first 60k iterations, and then decayed to $10^{-4}$ for the other 40k iterations. In finetuning stage, the learning rate is fixed to $10^{-4}$ for 20k iterations throughout. No extra data augmentation is used.

For the trade-off between efficiency and accuracy, we set $K=24$ and $M=32$ in out implement. Moreover, threshold $T$ is set to 0.1 in training for high recall and 0.5 in testing for high precision. Thanks to location by corners, the proposals are very accurate and only 200 proposals are used for further detection at test-time. The proposed method is implemented using Caffe \cite{CAFFE2014ACM}. All experiments are carried out on a standard PC with Intel i7-6800k CPU and a single NVIDIA 1080Ti GPU. All of our results are reported on single-scale testing images with a single model.

\subsection{Ablation Study}
To investigate the effectiveness of the proposed components, we conduct several ablation studies. We adopt Faster R-CNN \cite{FASTERRCNN2015NIPS} as our baseline model and convert the format of outputs to match the evaluation of ICDAR 2015. Results are shown in Table.~\ref{Table_1} and discussed in detail next. 

\setlength{\tabcolsep}{8pt}
\begin{table}[!htb]
\centering
\topcaption{\textbf{Ablations on ICDAR 2015.}}
\label{Table_1}
\begin{tabular}{|c|c|c|c|c|}
    \hline
    $Method$                & $Feature$ $Map$     & $Recall$    & $Precision$ & $F-measure$ \\\thickhline
    Baseline                & conv5, $stride=16$  & 0.606       & 0.553       & 0.579       \\\hline
    CRPN + RoI Pooling      & conv5, $stride=16$  & 0.707       & 0.611       & 0.656       \\\hline
    CRPN + RoI Pooling      & fusion, $stride=8$  & 0.767       & 0.781       & 0.774       \\\hline
    CRPN + RRoI Pooling     & fusion, $stride=8$  & 0.788       & 0.870       & 0.827       \\\hline
    CRPN + Dual-RoI Pooling & fusion, $stride=8$  & {\bf 0.807} & {\bf 0.887} & {\bf 0.845} \\\hline
\end{tabular}
\end{table}
\setlength{\tabcolsep}{1.4pt}

First, using Corner-based RPN instead of conventional RPN, the F-measure improves from 0.579 to 0.656. It is worth noting that the regression task in baseline is designed for refining the rectangular bounding-box. Next, with the aim to generate more accurate proposals, we predict on conv4 with fusion structure which is introduced in section 2.3, and the resulting model obtains a huge improvement in F-measure (0.656 vs 0.774). Replacing RoI Pooling with RRoI Pooling leads to an F-measure of 0.827, which also outperforms recent Rotation RPN (0.800 F-measure shown in Table.~\ref{Table_3}). From this we can conclude the Corner-based RPN is more effective to match the multi-oriented text in scene image. Finally, by incorporating with Dual-RoI Pooling, our method further obtains a gain of 1.8\% in F-measure and also achieves the highest recall and precision in the list. This suggest that the Dual-RoI Pooling is indeed helpful for the scene text detector.

\subsection{Experimental Results}
\begin{figure}[!htb]
\centering
\includegraphics[height=10.0cm]{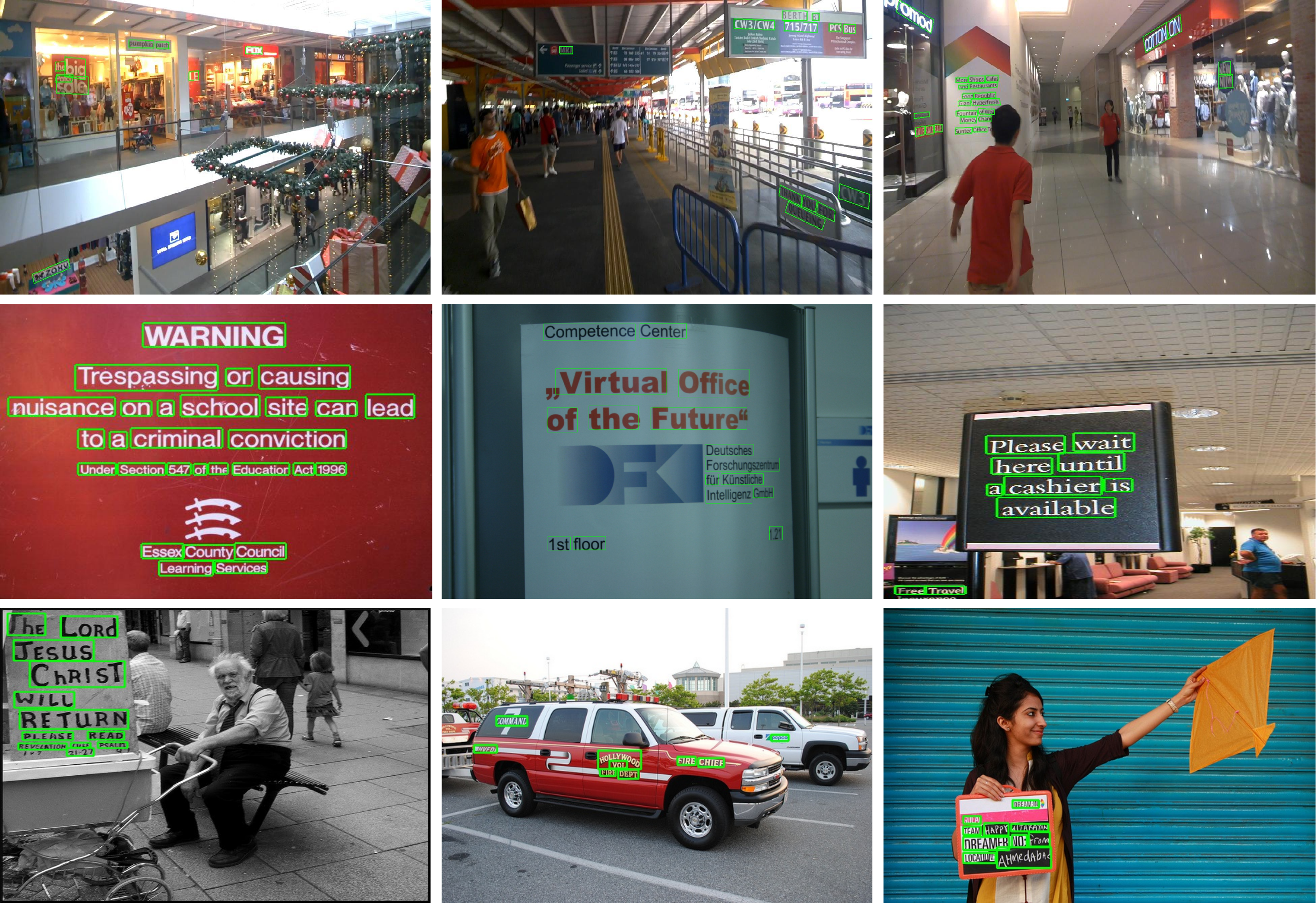}
\caption{\textbf{Selected results from the public benchmarks.} Viewing digitally with zoom is recommended.}
\label{Fig_5}
\end{figure}

We evaluate our method on public benchmarks, following the standard evaluation protocols in each field. Fig.\ref{Fig_5} shows some detection results from these datasets.

\subsubsection{ICDAR 2013 Focused Scene Text}

The ICDAR 2013 \cite{ICDAR2013} dataset consists of 229 training and 233 testing images which were captured by user explicitly detecting the focus of the camera on the text content of interest. It is the standard benchmark for evaluating horizontal or nearly horizontal text detection. In this benchmark, all of the testing images are resized with a fixed short side of 640 and we obtain an F-measure of 0.876 by using the ICDAR 2013 standard. As depicted in Table.~\ref{Table_2}, the result of proposed method is a comparable performance with a state-of-the-art method \cite{CTPN2016ECCV}, which utilizes a Connectionist Text Proposal Network to detect texts by predicting the sequence of fine-scale text components. Our method also outperforms other compared methods including DeepText \cite{DEEPTEXT2016ARXIV}, FCRN \cite{FCRN2016CVPR} and TextBoxes \cite{TEXTBOX2017AAAI}, which are mainly designed for nearly horizontal text detection. The proposed method runs at 9.1 fps, which is slightly faster than recent public work \cite{SSTD2017ICCV}. However, it is still too slow for real-time (25 fps) or nearly real-time application.

\setlength{\tabcolsep}{12pt}
\begin{table}[!htb]
\centering
\topcaption{\textbf{Results on ICDAR 2013 Focused Scene Text.} The results are reported in terms of Recall (R), Precision (P) and F-measure (F). (--) means no report in their papers.}
\label{Table_2}
\begin{tabular}{|c|c|c|c|c|c|c|c|} 
    \hline
    \multirow{2}*{$Method$} & \multicolumn{3}{c}{$ICDAR$ $Standard$} & \multicolumn{3}{|c|}{$DetEval$} & \multirow{2}*{$FPS$} \\
    \cline{2-7}
    ~                                 & $R$         & $P$         & $F$         & $R$         & $P$         & $F$         & ~          \\\thickhline
    FASText \cite{FASTEXT2015ICCV}    & 0.693	    & 0.840	      & 0.768	    & -- 	      & -- 	        & -- 	      & 6.7        \\\hline
    TextFlow \cite{TF2015ICCV}        & 0.759       & 0.852       & 0.803       & --          & --          & --          & 1.1        \\\hline
    TextBoxes \cite{TEXTBOX2017AAAI}  & 0.740       & 0.860       & 0.800       & 0.740       & 0.880       & 0.810       & 11.1       \\\hline
    FCRN \cite{FCRN2016CVPR}          & 0.764       & {\bf 0.938} & 0.842       & 0.755       & 0.920       & 0.830       & 0.8        \\\hline
    DeepText \cite{DEEPTEXT2016ARXIV} & 0.830       & 0.870       & 0.850       & --          & --          & --          & 0.6        \\\hline
    SegLink \cite{SEGLINK2017CVPR}    & --          & --          & --          & 0.830       & 0.877       & 0.853       & {\bf 20.6} \\\hline
    CTPN \cite{CTPN2016ECCV}          & 0.730       & 0.930       & 0.820       & 0.820       & {\bf 0.930} & {\bf 0.880} & 7.1        \\\hline
    SSTD \cite{SSTD2017ICCV}          & {\bf 0.860} & 0.880       & 0.870       & {\bf 0.860} & 0.890       & {\bf 0.880} & 7.7        \\\hline
    {\bf Ours}                        & 0.839       & 0.919       & {\bf 0.876} & 0.840       & 0.921       & 0.879       & 9.1        \\\hline
\end{tabular}
\end{table}
\setlength{\tabcolsep}{1.4pt}

\subsubsection{ICDAR 2015 Incidental Scene Text}

The ICDAR 2015 \cite{ICDAR2015} benchmark was released during the ICDAR 2015 Robust Reading Competition. It provides 1000 training and 500 test images which was collected without taking any specific prior attention. Therefore, it is more difficult than previous ICDAR challenges. In this benchmark, we rescale all of the testing images such that their short side is 900 pixels for detecting small text regions. As shown in Table.\ref{Table_3}, our method achieves an F-measure of 0.845, surpassing all of the anchor fashion methods, including RRPN \cite{RRPN2017ARXIV} and $R^2$CNN \cite{RRCNN2017ARXIV}, which are also extended from Faster R-CNN \cite{FASTERRCNN2015NIPS} framework and employ VGG-16 \cite{VGG2015ICLR} as the backbone of network. The FTSN \cite{FTSN2017ARXIV} based on detecting and segmenting text instance is very close to us in performance, but it is built on a more powerful network (ResNet-101 \cite{RESNET2016CVPR}). However, compared with EAST \cite{EAST2017CVPR} which utilizes a deep regression network that directly predicts text region with arbitrary orientations in full images, there is still a big gap in terms of recall.

\setlength{\tabcolsep}{14pt}
\begin{table}[!htb]
\centering
\topcaption{\textbf{Results on ICDAR 2015 Incidental Scene Text.}}
\label{Table_3}
\begin{tabular}{|c|c|c|c|} 
    \hline
    $Method$                            & $Recall$    & $Precision$ & $F-measure$ \\\thickhline
    MCLAB\_FCN \cite{MTD2016CVPR}       & 0.430       & 0.710       & 0.540       \\\hline
    CTPN \cite{CTPN2016ECCV}            & 0.520       & 0.740       & 0.610       \\\hline
    DMPNet \cite{DMPN2017CVPR}          & 0.680       & 0.730       & 0.710       \\\hline
    SegLink \cite{SEGLINK2017CVPR}      & 0.768       & 0.731       & 0.750       \\\hline
    SSTD \cite{SSTD2017ICCV}            & 0.730       & 0.800       & 0.770       \\\hline
    RRPN \cite{RRPN2017ARXIV}           & 0.770       & 0.840       & 0.800       \\\hline
    EAST \cite{EAST2017CVPR}            & {\bf 0.833} & 0.783       & 0.807       \\\hline
    NLPR\_CASIA \cite{DDR2017ICCV}      & 0.800       & 0.820       & 0.810       \\\hline
    R{$^{2}$}CNN \cite{RRCNN2017ARXIV}  & 0.797       & 0.856       & 0.825       \\\hline
    FTSN \cite{FTSN2017ARXIV}           & 0.800       & 0.886       & 0.841       \\\hline
    \textbf{Ours}                       & 0.807       & {\bf 0.887} & {\bf 0.845} \\\hline
\end{tabular}
\end{table}
\setlength{\tabcolsep}{1.4pt}

\subsubsection{COCO-Text}

The original images of COCO-Text \cite{COCO2016ARXIV} are harvested from Microsoft COCO \cite{MSCOCO2014ECCV} dataset, and it contains 43686 training images and 20000 images for validation and testing. It is the largest dataset for text detection and recognition in scene images to date. The testing images are resized with a fixed short side of 900, and our method achieves 0.633, 0.555 and 0.591 in recall, precision and F-measure by using the online evaluation system provided officially, as shown in Table.\ref{Table_4}. It is worth noting that no images from COCO-Text are involved in training phase. The presented results demonstrate that our method is capable of applying practically in the unseen contexts. 

\setlength{\tabcolsep}{14pt}
\begin{table}[!htb]
\centering
\topcaption{\textbf{Results on COCO-Text}. The results of compared methods are grasped from the public COCO-Text leaderboard.}
\label{Table_4}
\begin{tabular}{|c|c|c|c|}
    \hline
    $Method$            & $Recall$    & $Precision$ & $F-measure$ \\\thickhline
    SCUT\_DLVClab       & 0.625       & 0.316       & 0.420       \\\hline
    SARI\_FDU\_RRPN     & 0.632       & 0.333       & 0.436       \\\hline
    UM                  & {\bf 0.654} & 0.475       & 0.551       \\\hline
    TDN\_SJTU\_v2       & 0.543       & {\bf 0.624} & 0.580       \\\hline
    Text\_Detection\_DL & 0.618       & 0.609       & {\bf 0.613} \\\hline
    \textbf{Ours}       & 0.633       & 0.555       & 0.591       \\\hline
\end{tabular}
\end{table}
\setlength{\tabcolsep}{1.4pt}

\subsection{Limitations}
\begin{figure}[!htb]
  \centering
  \subfigure[]{
    \label{Fig_6_a} 
    \includegraphics[height=3.5cm]{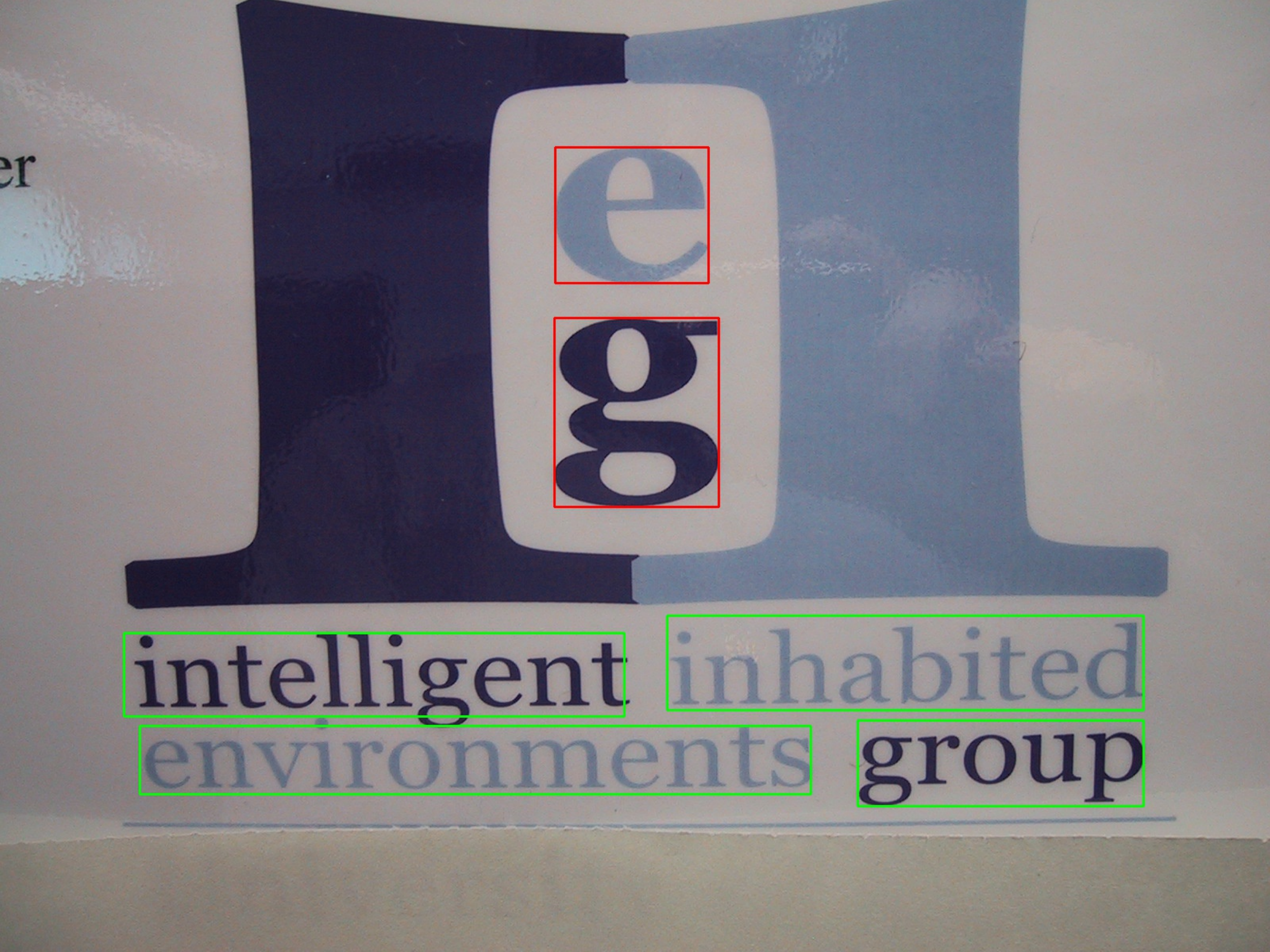}}
  \hspace{-0.5ex}
  \subfigure[]{
    \label{Fig_6_b} 
    \includegraphics[height=3.5cm]{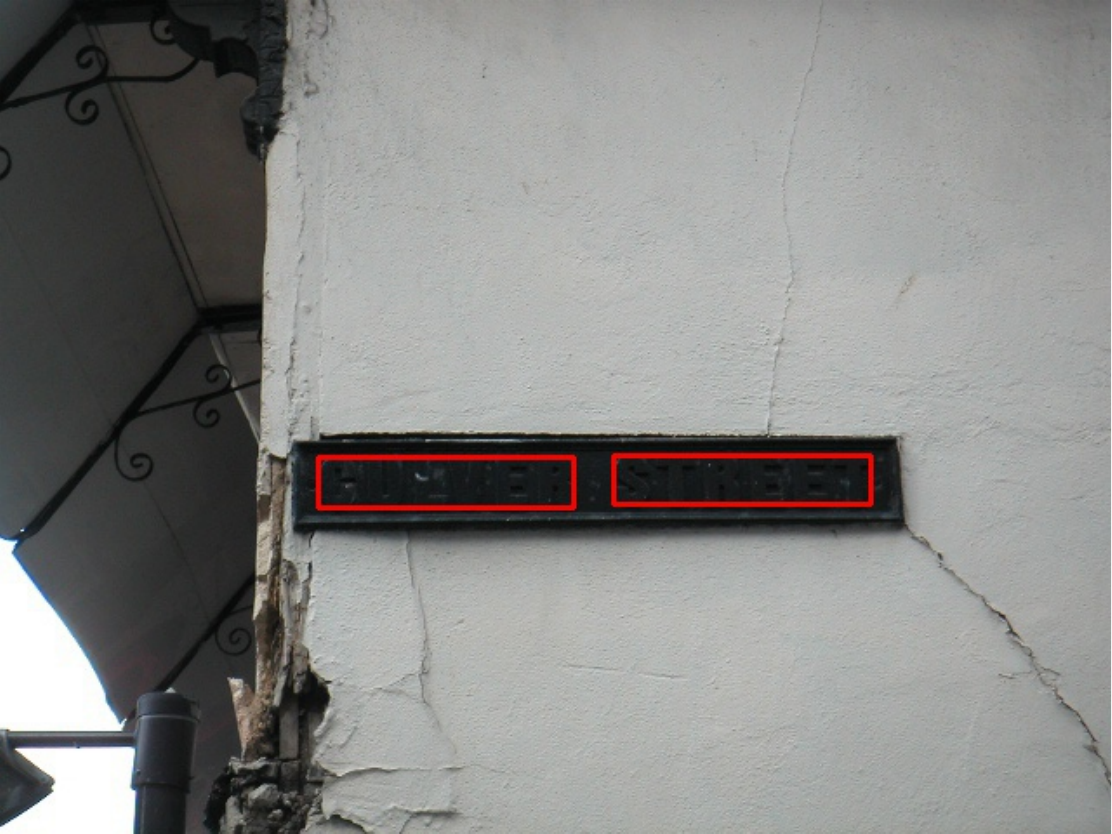}}
  \hspace{-0.5ex}
  \subfigure[]{
    \label{Fig_6_c} 
    \includegraphics[height=3.5cm]{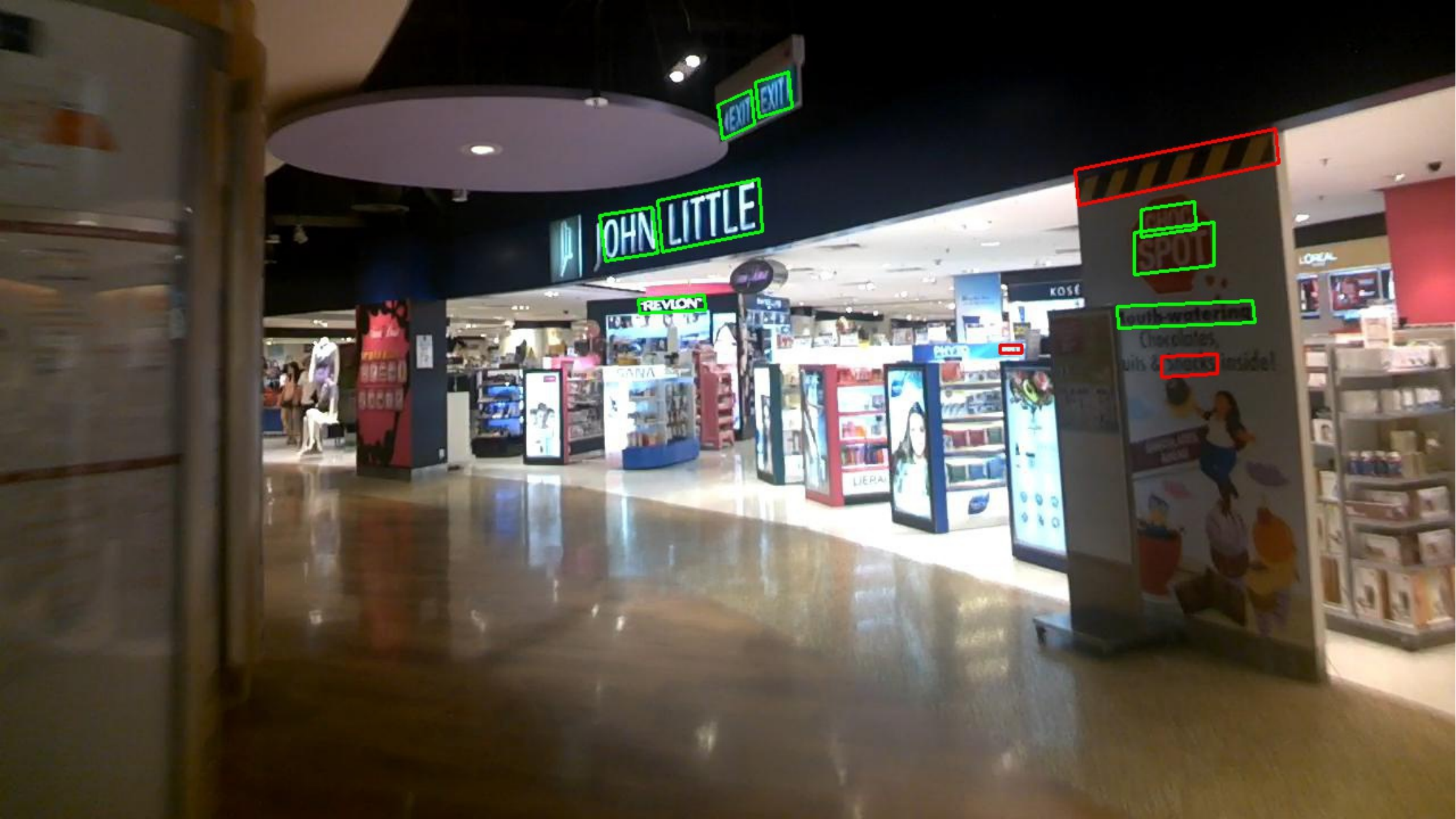}}
  \caption{\textbf{Failure cases on ICDAR 2013 and 2015.} Red boxes are false positives or true negatives. Best viewed zoomed in color.}
  \label{Fig_6} 
\end{figure}

We further analyze the limitations of our method. In essence, the proposed framework utilizes a bottom-up strategy from corners to quadrilateral bounding-box. Hence the problem of accumulation of errors still exists. As shown in Fig.\ref{Fig_6}.a, our method may fail to detect text with single character because we drop it in training stage for high precision. Moreover, our method still struggles with some extreme scenarios, such as the examples in Fig.\ref{Fig_6}.b, c. Finally, we have to magnify the image size in order to detect small text region, which also limits the efficiency of our method.

\section{Conclusion and Future Work}
This paper presents an intuitive region-based method towards multi-oriented text detection by learning the idea from the process of object annotation. We discard the anchor strategy and employ corners to estimate the possible locations of text instances. Based on corners, our method is flexible to generate quadrilateral proposals for capturing various kinds of texts. Moreover, we design a built-in data augmentation module inside the region-wise subnetwork, which not only utilizes training data more efficiently, but also improves the robustness of the resulting model. In the future, the performance of our method can be further improved by using much stronger networks, such as ResNet \cite{RESNET2016CVPR} or DenseNet \cite{DENSENET2017CVPR}. Additionally, we are also interested in extending this method to an end-to-end text reading system.

\section*{References}

\bibliography{mybibfile}

\end{document}